\documentclass[10pt,twocolumn,letterpaper]{article}

\usepackage{iccv}
\usepackage{times}
\usepackage{epsfig}
\usepackage{graphicx}
\usepackage{amsmath}
\usepackage{amssymb}

\usepackage{caption}
\usepackage{comment}
\usepackage{makecell}
\usepackage{array}
\usepackage[breaklinks=true,bookmarks=false]{hyperref}

\iccvfinalcopy 

\begin{document}

\title{Transferring Knowledge with Attention Distillation for Multi-Domain Image-to-Image Translation}

\author{Runze Li\thanks{Runze Li and Tomaso Fontanini contributed equally to
this work.}\\
University of California Riverside\\
{\tt\small runze.li@email.ucr.edu}
\and
Tomaso Fontanini\footnotemark[1]\\
University of Parma\\
{\tt\small tomaso.fontanini@studenti.unipr.it}

\and
Luca Donati \\
University of Parma\\
{\tt\small luca.donati@unipr.it}

\and
Andrea Prati\\
University of Parma\\
{\tt\small andrea.prati@unipr.it}

\and
Bir Bhanu\\
University of California Riverside\\
{\tt\small bhanu@ee.ucr.edu}
}

\maketitle

\begin{abstract}
    Gradient-based attention modeling has been used widely as a way to visualize and understand convolutional neural networks. However, exploiting  these visual explanations during the training of generative adversarial networks (GANs) is an unexplored area in computer vision research. Indeed, we argue that this kind of information can be used to influence GANs training in a positive way. For this reason, in this paper, it is shown how gradient-based attentions can be used as knowledge to be conveyed in a teacher-student paradigm for multi-domain image-to-image translation tasks in order to improve the results of the student architecture. Further, it is demonstrated how ``pseudo"-attentions can also be employed during training when teacher and student networks are trained on different domains which share some similarities. The approach is validated on multi-domain facial attributes transfer and human expression synthesis showing both qualitative and quantitative results. Code in the supplementary material will be released publicly.
\end{abstract}

\begin{figure*}[t!]
\begin{center}
 \includegraphics[width=1.0\textwidth]{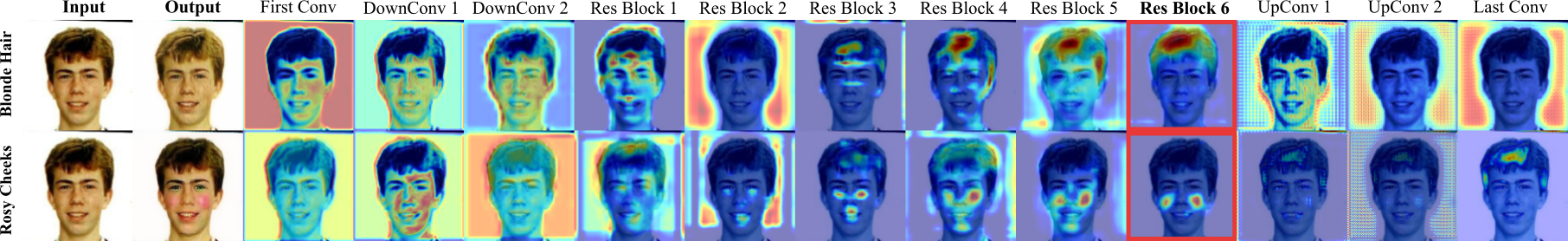}
\end{center}
 \captionof{figure}{Visual explanations extracted from each convolutional/residual block over the conditional GAN in the image-to-image translation task using attention maps. The input to the model is sources (\eg{} face) images with target domains (\eg{} facial attributes) and the output is the transferred images where target domains are expected to be applied on. In the first row the target domain is \textit{\textbf{blonde hair}} while in the second row is \textit{\textbf{rosy cheeks}}. The most refined attention can be seen in the last residual block (highlighted in red). In the attention map, the more red the region is, the more important it is to the network.}  \label{fig:generator_maps}
\end{figure*}

\section{Introduction}

Recent progress in understanding and explaining neural networks has led to many applications in real-world tasks. Inspired by Zeiler and Fergus \cite{zeiler2014visualizing}, much effort has been dedicated in visualizing and understanding feature activations in convolutional neural networks (CNNs) by generating attention maps that highlight regions which are considered to be important for the network goals. Given a trained CNN model, attention maps indicate where an object is located in an image, \eg{}, an airplane is in the top area of the image. This kind of visualization using attention maps helps to explain why this image is classified to the \textit{airplane} class. 
Following this line of research, Wenqian \etal{} \cite{liu2020towards} proposed a technique to visually explain Variational Autoencoders (VAEs), using gradient-based attention maps, showing the capability of attention maps in understanding generative models by means of visual explanations.

Exploiting the advances in model explainability, efforts were made in exploring how to utilize valuable information contained in attention maps in order to improve the performance of CNNs. Li \etal{} \cite{Li_2018_CVPR} used generated attention maps as visual guidance for training CNNs and observed improvements in model generalizability in traditional image classification and segmentation tasks. Zagoruyko \etal \cite{zagoruyko2016paying} built a system to transfer attentions from a teacher network to a smaller student network and observed a better performance of the student network in terms of image classification performance. Dhar \etal \cite{Dhar_2019_CVPR} proposed an approach with attention distillation loss for incremental learning. In spite of these significant advances, transferring knowledge to Generative Adversarial Networks (GANs) by using attention maps is an area that has not been explored yet. 

With the introduction of GANs and its many variants, image generation has made a gigantic leap forward, especially in terms of photorealism. 
Still, often it is necessary to condition the generative model in order to have control over its output. This is the case of image-to-image translation, where the objective is to have a more useful representation of the input data that can be later used for several tasks. 

In this paper, we argue that exploiting visual explanations in image-to-image conditional adversarial networks is a fundamental step in order to improve upon them. Following the work of \cite{selvaraju2017grad}, we are able to generate attention maps in the generator of a multi-domain image-to-image conditional GAN with only a small change in the model.
Multi-domain image-to-image translation task requires the system to translate a single input image into multiple domains (\eg{}, facial attributes). 
The produced attention maps highlight the features where the network is focusing for a certain domain and also allow to identify which layers in the generator are more devoted to the domain translation task. An example is presented in Fig. \ref{fig:generator_maps}. Once we have this information, it is possible to use the attention maps to convey knowledge to another network that shares the same image-to-image translation task by following a teacher-student paradigm, thus improving its generated samples. To this end, we present a learning objective that uses attention distillation loss and show how this loss term can be integrated in the student training. In other words, the teacher network can provide a meaningful visual explanation for the translation task, in order to guide the training of the student network. In addition to that, another possible application is to use what we call ``pseudo"-attention maps, that are attention maps generated from a set of domains and used as supervisions for a different set of domains in order to help the generator to produce better samples belonging to this new set of domains.


The main contributions of this work are as follows:
\begin{itemize}
    \item A system that identifies visual explanations for multi-domain image-to-image translation conditional GANs by adapting gradient-based attentions, it demonstrates how this kind of information visualizes the domain transfer task and can be used for an improved training. 
    \item A system consisting of a teacher and a student network, where the teacher uses only the attention generated in its latent space to improve the performance of the student network. The system forces the student model to ``look'' at the images in the same way as the teacher model, directing the student training to be as close as possible to the one of the teacher.
    \item A system that 
    utilizes ``pseudo"-attention maps generated from a set of domains from the teacher model 
    to help the student model to learn a set of different domains.
\end{itemize}


\section{Related Work}
\noindent\textbf{Conditional GANs for image-to-image translation.} Generative Adversarial Networks (GANs) \cite{goodfellow2014generative} in their many variations represent the state-of-the-art for photo-realistic image synthesis today. In particular, when a much finer control over the output is required, conditional GANs (cGANs) \cite{mirza2014conditional} allow the generation of images from text \cite{reed2016generative, zhang2018stackgan++}, class labels \cite{brock2018large}, sketches or textures \cite{sangkloy2017scribbler,xian2018texturegan}. Furthermore, while initially a paired dataset was required \cite{liu2017unsupervised}, CycleGAN  \cite{zhu2017unpaired} proved that a conditional GAN can be successfully trained in an unpaired way. Another relevant feature that most cGANs lack is the ability of producing images belonging to different classes or domains using a single architecture. Some models \cite{huang2018multimodal,liu2019few} achieve that by using adaptive instance normalization layers \cite{huang2017arbitrary} combined with a class-specific encoder and a content-specific encoder. On the other hand, StarGAN \cite{choi2018stargan} and its variants \cite{siddiquee2019learning,wu2019relgan} take as input both an image and the target domain label learning to flexibly execute the translation using only one underlying representation. Finally, cGANs can also be combined with meta-learning for greater flexibility and robustness \cite{fontanini2020metalgan}.

\noindent\textbf{CNNs visual attention explanation.} Deep Convolutional Networks have achieved astounding results in most computer vision tasks, but they are still mainly used as black boxes. For this reason, explaining their behaviour by visualizing `where they look' when making a decision has attracted lots of interest in the last years. Particularly, after the initial work of \cite{mahendran2015understanding} and \cite{zeiler2014visualizing}, Zhou \etal{} \cite{zhou2016learning} provided a method for  generating \textit{class activation mappings} (CAM) by using the global average pooling. The main drawback of this solution is that it is applicable only to a specific set of CNNs, that is the ones having the features maps directly preceding softmax layers.

For this reason, Grad-CAM \cite{selvaraju2017grad} and its variant Grad-CAM++ \cite{chattopadhay2018grad} were proposed. They both use gradient of the score for a class $c$ with respect to a specific feature maps $F_k$ to obtain the class-discriminative localization map and, thus, can be considered gradient-based. Moreover, they are applicable to a very wide range of architectures without requiring any structural change in the network, unlike response-based approaches \cite{fukui2019attention, zhang2018top, zhou2016learning} that instead use additional trainable units. Recently, the concept of visual attention was also extended to GANs \cite{zhang2019self} and VAE \cite{liu2020towards}.

\begin{figure*}[t!]
\begin{center}
\includegraphics[width=\textwidth]{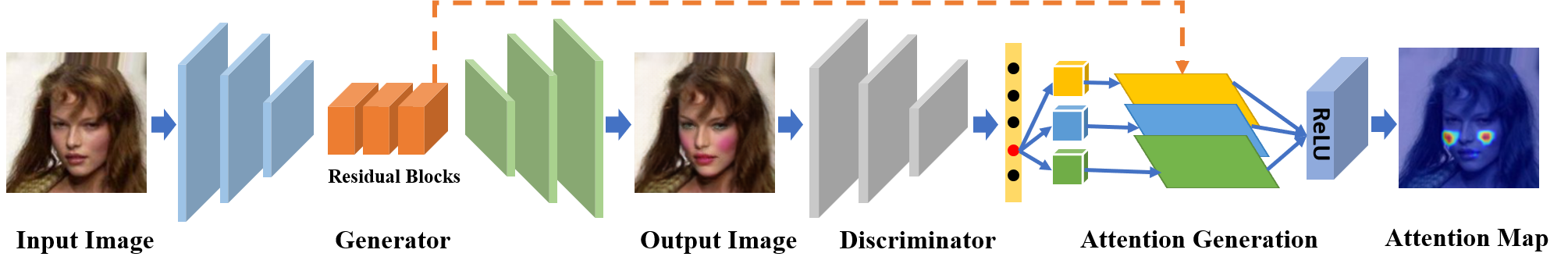}
\end{center}
\vspace{-2mm}
   \captionof{figure}{Attention generation with image-to-image conditional GAN. 
   }
\label{fig:attention_pipeline}
\vspace{-2mm}
\end{figure*}

\noindent\textbf{Knowledge distillation in neural networks.} Knowledge distillation involves transferring knowledge from a more complex network (teacher) to a simpler and lighter network (student) sharing the same task \cite{hinton2015distilling}. The goal is to have the student to reach almost the same results as the teacher. Many techniques have been developed in this area \cite{chen2017learning,park2019relational,polino2018model}, with \cite{zagoruyko2016paying} being the first to use attention transfer to improve the performance of a student classification network. Previous methods were almost entirely applied over recongnition or classification models, while \cite{aguinaldo2019compressing} introduced a method working on unconditional GANs. Recently, Li \textit{et al.} \cite{li2020gan} proposed a compression algorithm for cGANs using feature distillation and neural architecture search.

\section{Technical Approach}
As the case study in our paper, we take a conditional GAN for multi-domain image-to-image translation, but the attention map generation pipeline 
can be applied in a wider variety of GAN networks in order to explore explainabilities of generative models.

Some examples of the generated attention maps 
are shown in Fig. \ref{fig:generator_maps}. We observe that those facial attributes used as target domains 
are highlighted as regions in attention maps. For example, \textit{\textbf{cheekbones}} of the face are highlighted as ``red" in the attention map (\textbf{Res Block 6}) in Fig. \ref{fig:generator_maps}, which means that the network is focusing on those regions when applying the target domain \textit{\textbf{rosy cheeks}} to obtain the output images (\textbf{Output}).

\subsection{Generate Explanations for Image-to-Image Translation Conditional GANs}
\label{section3_1}

Adapting the framework of Grad-CAM \cite{selvaraju2017grad}, we obtain an attention map corresponding to the input sample from our image-to-image conditional GAN model that has been given an input image \textit{x} and a set of target domains \textbf{\textit{c}}. For each class \textit{c}, from the ground-truth labels of target domains, we compute the gradient of score $y^c$ corresponding to the class \textit{c}. We backpropagate the gradients directly from the classification output of the discriminator to the convolutional layers of the generator with feature maps $\mathbf{F} \in \mathbb{R}^{n\times h\times w}$, with $n$, $h$ and $w$ being depth, height and width of the feature map, respectively, obtaining attention maps $\mathbf{A}^c$ corresponding to $y^c$. Specifically, we calculate $\mathbf{A}^c$ by using the following equation:
\begin{equation}
\mathbf{A}^c=ReLU\left(\sum_{k=1}^{n}\alpha_{k}^c\mathbf{F}_{k}\right)
\label{eq:att_zi}
\end{equation} 
where the scalar $\alpha_{k}^c=\text{GAP}\left(\frac{\partial y^c}{\partial \mathbf{F}_{k}}\right)$ and $\mathbf{F}_{k}$ is the $k^{th}$ feature channel ($k=1,\ldots,n$) of the feature maps $\mathbf{F}$, with $\frac{\partial y^c}{\partial \mathbf{F}^{k}}$ representing the gradient of the score $y^c$ with respect to the feature maps $\mathbf{F}^k$. The global average pooling (GAP) operation is used to obtain scalar $\alpha_{k}^c$ as:
\begin{equation}
\alpha_{k}^c=\frac{1}{S}\sum_{m=1}^{h}\sum_{n=1}^{w}\left(\frac{\partial y^c}{\partial F^{mn}_{k}}\right)
\end{equation}
where $S=h\times w$ and $F^{mn}_{k}$ is the pixel value at location $(m,n)$ of the $h\times w$ matrix $\mathbf{F}_{k}$. The attention map generation process is illustrated in Fig. \ref{fig:attention_pipeline}.

\subsection{Network Architecture}
\label{section3_2}
\begin{figure*}[ht]
\begin{center}
\includegraphics[width=0.8\linewidth]{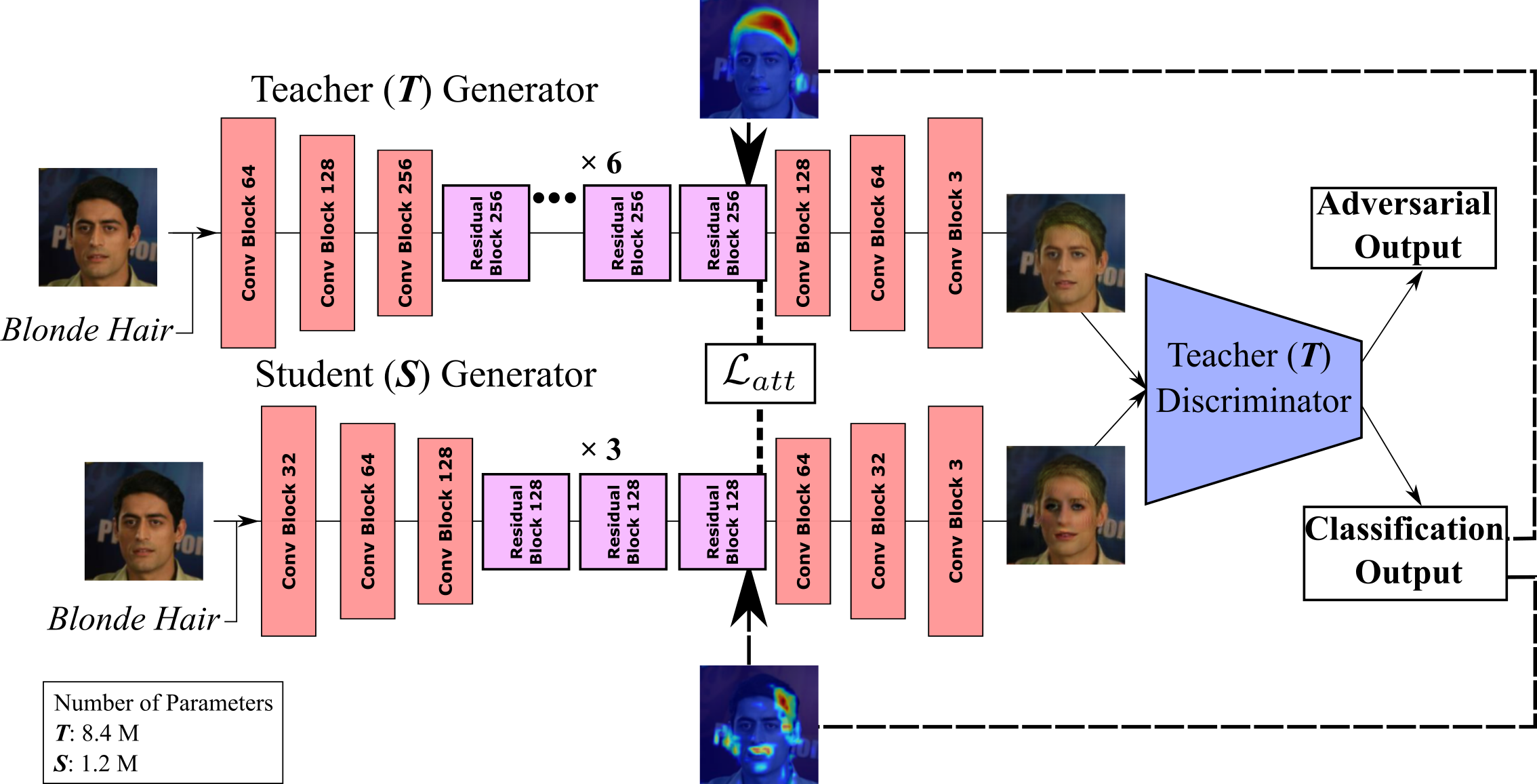}
\end{center}
   \caption{Summary of the full system architecture: the \textit{Teacher} \textbf{\textit{T}} network represents the full model with 6 residual blocks while the \textit{Student} \textbf{\textit{S}} network is a smaller and compressed version of \textbf{\textit{T}} model with half the number of features maps in each layer of the generator and 3 residual blocks. During training, our proposed attention distillation loss $\mathcal{L}_{att}$ is calculated using the attention maps obtained from the teacher and the student generators using the same input image and target class and backpropagating the classification output of the teacher discriminator. In addition to that, in Sec. \ref{sec_4_3}, ablation experiments are performed using an ever smaller student called \textit{\textbf{S-Lite}} with just 0.16 million parameters.}
\label{fig:networks_arch}
\end{figure*}

Our system is composed by a teacher \textbf{\textit{T}} and a student \textbf{\textit{S}} network. Both of them are multi-domain image-to-image conditional GANs consisting of a generator and a discriminator conditioned by the label of the class that we want to transfer over the input image. The generator takes as input an image and the target class and returns the translated image, while the discriminator takes the translated image as input and returns an adversarial output and a domain classification output. For the network architectures, we draw inspiration from StarGAN \cite{choi2018stargan} and use it as the teacher \textit{\textbf{T}} network. In particular, StarGAN generator is composed by an encoder and a decoder. The encoder consists of a series of convolutional blocks, generating feature maps, followed by a group of residual blocks (6 in the teacher model), and the decoder is composed by a number of consecutive deconvolutional blocks and outputs the transferred images.



In the student \textbf{\textit{S}} network, 
we reduced feature map dimensions and the number of residual block since doing so has a significant effect on the complexity of the network: the number of feature maps in each layer of the student generator is half that of each layer of the teacher generator  
and there are only 3 residual blocks instead of 6. In additional to that, we also design \textbf{\textit{S-Lite}} which is an even lighter student network whose details are described in Sec. \ref{sec_4_3}. On the other side, the discriminator is the same for both student and teacher models. 
Teacher \textbf{\textit{T}} and student \textbf{\textit{S}} generators have 8.4 million and 1.2 million parameters, respectively.

    





\subsection{Attention Knowledge Transfer}
\label{section_3_3}
In this section, we present our approach for conveying knowledge via attention distillation between the teacher and the student network. We produce visual explanations over the generator of the teacher network using the method introduced in Sec. \ref{section3_1}. Analyzing these attention maps, like the ones in Fig. \ref{fig:generator_maps}, it is clear how the better visual explanation for the image-to-image translation task can be found in the last residual block, while no useful information can be extracted from the downsampling and upsampling layers. 
This implies that we should direct the knowledge transfer by deriving an attention distillation loss over the maps extracted from 
the last residual block, as opposed to extracting them from each layer of the teacher generator, thus, reducing the training complexity by a considerable amount.

\subsubsection{Training with Attention Knowledge Transfer}

In order to train the full system, we defined a set of losses. We use adversarial loss $\mathcal{L}_{adv}$, reconstruction loss $\mathcal{L}_{rec}$ and domain classification loss  $\mathcal{L}_{cls}$ as \cite{choi2018stargan} for pre-training the teacher \textbf{\textit{T}} network. Furthermore, we use the same losses to train the student \textbf{\textit{S}} network with the addition of an attention distillation loss $\mathcal{L}_{att}$ for transferring the knowledge encoded in visual explanations from the teacher to the student. 

The attention distillation loss is defined as:

\begin{equation}
    \mathcal{L}_{att} = \mathbb{E}_{x,c}\left[\parallel \mathbf{A}^{c}_{\boldsymbol{\mathit{T}}}(x) - \mathbf{A}^{c}_{\boldsymbol{\mathit{S}}}(x) \parallel\right]
\end{equation}

\noindent where $\mathbf{A}^{c}_{\boldsymbol{\mathit{T}}}(x)$ and $\mathbf{A}^{c}_{\boldsymbol{\mathit{S}}}(x)$ are the attention maps calculated over the last convolution of the last residual block of the teacher and student generator, respectively, using the same input image $x$ and the same target domain $c$.





During each training step of the student network, we trained the discriminator to distinguish between real and fake samples and to correctly classify the images into the multiple domains. The generator 
is trained to fool the discriminator by producing better samples belonging to the  target domains. The reconstruction loss serves the purpose of maintaining the content of the input image even after the translation, \eg{} applying mustache to a person must not change the subject's identity.


In addition, we applied the attention distillation loss to push the results from the student network to be as close as possible to the ones from the teacher network (Fig. \ref{fig:networks_arch}).

Finally, the full student training objective is:
\vspace{-2mm}
\begin{equation}
    \mathcal{L}_{D}^{\boldsymbol{\mathit{S}}} = \mathcal{L}_{adv} + \lambda_{cls}\mathcal{L}_{cls}
\end{equation}
\vspace{-5mm}
\begin{equation}
    \mathcal{L}_{G}^{\boldsymbol{\mathit{S}}} = \mathcal{L}_{adv} + \lambda_{cls}\mathcal{L}_{cls} + \lambda_{rec}\mathcal{L}_{rec} + \lambda_{att}\mathcal{L}_{att}
\end{equation}

\noindent for the discriminator and the generator, respectively. We use $\lambda_{cls} = 1$, $\lambda_{rec} = 10$ and $\lambda_{att} = 10$ in our experiments.

\subsection{Pseudo-Attention for Knowledge Transfer}
\label{section3_4}
In this section, we will discuss how to transfer knowledge from a teacher \textbf{\textit{T}} network trained on a set of domains to a student \textbf{\textit{S}} network that employs a \textbf{\textit{new}} set of \textbf{\textit{different}} domains. Our intuition is that when the input images are translated to a different target domain, the modified areas in the samples might be shared through different domains. For example, in our task 
in order to translate an input image to the output images with attributes \textit{\textbf{black hair}}, \textbf{\textit{blonde hair}} or \textbf{\textit{brown hair}}, the network is supposed to mainly pay attention to the region 
corresponding to the attribute  \textit{``hair"} in the input image. Given these observations, in this section, we start from defining the ``pseudo"-attention maps and present how to derive attention distillation loss by using the ``pseudo"-attention maps for conveying knowledge in our image-to-image translation task. 

\subsubsection{Training with Pseudo-Attention Knowledge Transfer}
As introduced in Sec. \ref{section3_2}, we employed a teacher-student paradigm in order to transfer knowledge using attention maps on a multi-domain image-to-image translation task. 
Firstly, given a set of target domains \textbf{\textit{c}}, we selected a 
second set \textbf{\textit{c}}$^{pse}$ where 
a domain \textit{c}$^{pse}$, indicated by a certain attribute, would share common regions in the images with 
a domain \textit{c}. Next, we fully trained the teacher \textbf{\textit{T}} network 
on target domains \textbf{\textit{c}}$^{pse}$ in order to calculate attention maps from it. Finally, we started training our teacher-student system, where the student \textbf{\textit{S}} network is trained 
on target domains \textbf{\textit{c}}. While training the student network, we generated attention maps and calculated the attention distillation loss to transfer knowledge from the teacher network to the student network. More in detail, given an input image $x$ and a target domain $c$, we calculated the attention maps using the corresponding domain \textit{$c^{pse}$} for both the teacher and the student networks. 
Indeed, attention maps $\mathbf{A}^{c^{pse}}_{\boldsymbol{\mathit{T}}}$ and $\mathbf{A}^{c^{pse}}_{\boldsymbol{\mathit{S}}}$ are obtained  backpropagating the class score $y^{c^{pse}}$ from the discriminator to the last convolution of the last residual block of the teacher and student generators, respectively.
Since the student network is trained to translate input images to a set of target domains \textbf{\textit{c}}, which are different domains from \textbf{\textit{c}}$^{pse}$, we call the attentions obtained in this way ``pseudo"-attentions. Thus, the attention distillation loss with ``pseudo"-attentions is calculated as:

\begin{equation}
    \mathcal{L}^{pse}_{att} = \mathbb{E}_{x,c}[\parallel \mathbf{A}^{c^{pse}}_{\boldsymbol{\mathit{T}}}(x) - \mathbf{A}^{c^{pse}}_{\boldsymbol{\mathit{S}}}(x) \parallel]
\end{equation}

Our intuition with the ``pseudo"-attention distillation loss is to use one pre-trained teacher network for different student networks targeting on different domains that share some common regions in the images with the set of domains used in the teacher network.


The training objective of discriminator is the same as Equation 4 and the objective of generator in student network is:
\begin{equation}
    \mathcal{L}_{G}^{\boldsymbol{\mathit{S}}} = \mathcal{L}_{adv} + \lambda_{cls}\mathcal{L}_{cls} + \lambda_{rec}\mathcal{L}_{rec} + \lambda_{att}\mathcal{L}^{pse}_{att}
\end{equation}

\section{Experiments}

In this section we has shown results using two different experimental settings: (a) teacher and student networks were trained using the same set of domains and the attention was used to make the student mimic the teacher behaviour as defined in Section \ref{section_3_3}, (b) the teacher was trained on a set of domains and the student on a set of different domains and the ``pseudo"-attention was used as guidance to help the student to learn the new set of domains as defined in Section \ref{section3_4}. For each experiment we have shown both the qualitative and quantitative evaluation using FID score \cite{heusel2017gans} to measure the overall visual quality of the samples and classification accuracy to determine the quality of the domain translation. 

The core experiments were performed using the CelebA dataset \cite{liu2015faceattributes}, which is a large-scale facial attributes dataset with more than 200k images and 40 attributes. Facial attributes were also very suited to show the benefits of our system, since, in order to correctly transfer them, the network needs to learn their location on the human face and visual attention can catch this kind of information. In addition, we have conducted ablation experiments on RaFD dataset \cite{langner2010presentation} which includes 8k images distributed in 8 emotional expressions. 

\subsection{Attention Knowledge Transfer}
\label{section4_1}


\begin{figure*}[ht]
\begin{center}
\includegraphics[width=1.0\linewidth]{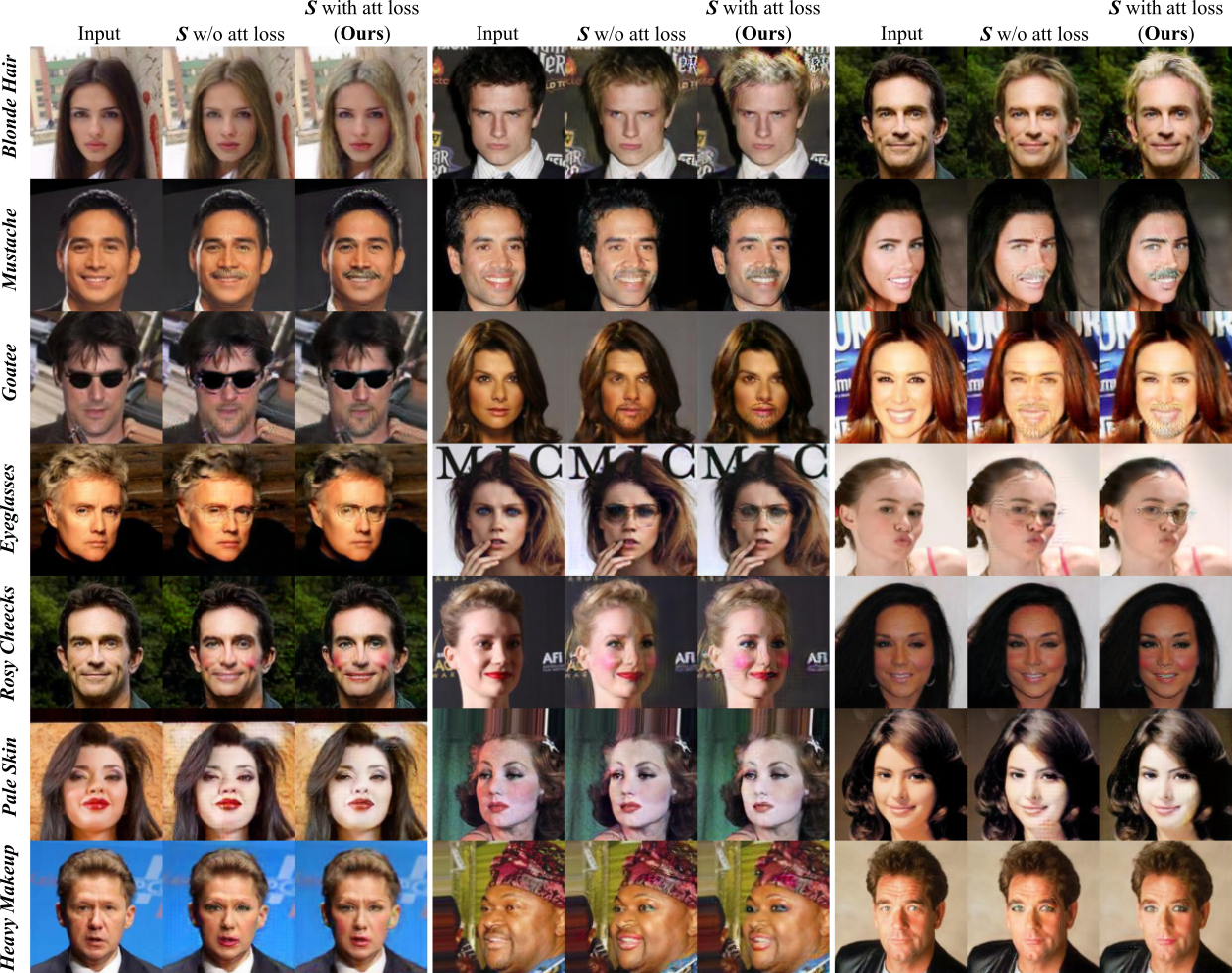}
\end{center}
   \caption{Collection of comparisons between results obtained from the student \textit{\textbf{S}} without using the attention distillation loss and results obtained when training \textit{\textbf{S}} with the attention distillation loss proposed in this paper. Each triplet images consist of the input image  with target domains, the output image from the student network and the output from the student network trained using our proposed method.}  
\label{fig:qualitative_res}
\end{figure*}

In this set of experiments the objective is to use attention to transfer knowledge between a teacher \textbf{\textit{T}} network and a smaller student \textbf{\textit{S}} network. We used teacher and student architectures from Fig. \ref{fig:networks_arch}, respectively. The models were trained to translate 7 different facial attributes that are \textit{\textbf{goatee}}, \textit{\textbf{rosy cheeks}}, \textit{\textbf{eyeglasses}}, \textit{\textbf{mustache}}, \textit{\textbf{blond hair}}, \textit{\textbf{pale skin}} and \textit{\textbf{heavy makeup}} and we transferred knowledge conveyed via visual attentions from the last residual block of \textbf{\textit{T}} generator to the last residual block of the \textbf{\textit{S}} generator. 

\begin{figure}[ht]
\begin{center}
\includegraphics[width=\linewidth]{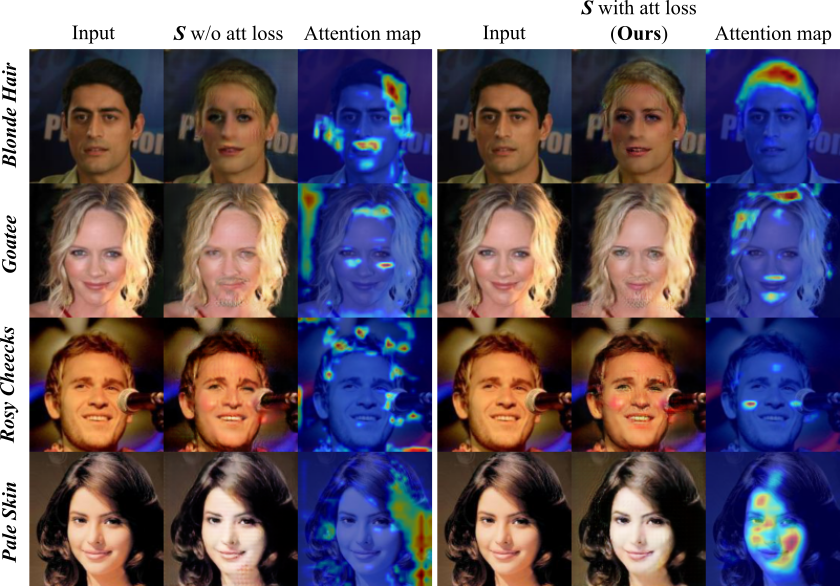}
\end{center}
   \vspace{-2mm}
   \caption{Results (translated images and attention maps) without attention loss (left) and with attention loss (right) in generated from student network \textbf{\textit{S}}.} 
\label{fig:results_small_stud}
\vspace{-2mm}
\end{figure}

\begin{table*}[ht]
    \centering
    \resizebox{\textwidth}{!}{
    \begin{tabular}{|c|c|c|c|c|c|c|c||c||c|}
    \hline
         & Blonde Hair $\uparrow$ & Mustache $\uparrow$ & Goatee $\uparrow$ & Eyeglasses $\uparrow$ & Rosy Cheeks $\uparrow$ & Pale Skin $\uparrow$ & Heavy Makeup $\uparrow$ & Mean $\uparrow$ & FID $\downarrow$ \\ \hline
          Stargan \cite{choi2018stargan} (upper bound) & 83.72\% & 28.05\% & 55.36\% & 73.25\% & 40.34\% & 78.36\% & 96.54\% & 65.08\% & 18.61 \\ \hline \hline
        \textbf{\textit{S}} w/o att loss & 75.46\% & 24.71\% & \textbf{51.38\%} & 56.48\% & 30.59\% & 67.14\% & 94.78\% & 57.22\% & 21.48 \\ \hline
         \textbf{\textit{S} with att loss (Ours)} & \textbf{77.81\%} & \textbf{24.87\%} & 48.46\% & \textbf{62.65\%} & \textbf{35.40\%} & \textbf{70.90\%} & \textbf{96.25\%} & \textbf{59.47\%} & \textbf{21.02}\\ 
         \hline
    \end{tabular}
    }
    \caption{Classification results and FID scores over 7 different facial attributes. First row is the StarGAN model (acting as upper bound as it is a full model). 
    The bold results represent the best results between the student network \textbf{\textit{S}} trained without and with attention distillation loss, respectively.}
    \label{tab:class_stud}
\end{table*}

\noindent\textbf{Qualitative Results.} Qualitative results are presented in Fig. \ref{fig:qualitative_res}. Thanks to the contribution of the proposed attention distillation loss, target domains are applied much more strongly to the input images. This is particularly true for facial attributes \textit{blonde hair}, \textit{rosy checks}, \textit{mustache} and \textit{pale skin} whose application was unsatisfactory in the student without attention loss. Moreover, target facial attributes of \textit{eyeglasses} and \textit{heavy makeup} are applied more convincingly to the input images. Finally, some undesired changes that can happen during the translation of certain domains are less frequent. More specifically, when translating some facial attributes related to one particular gender, \ie{} \textit{goatee}, it can happen that gender is also translated to the output image. For example, in the third row of the Fig. \ref{fig:qualitative_res}, the facial attribute \textit{goatee} is applied to two female face images and indeed without attention loss their appearance resembles the one of a man much more than when attention loss is applied. 

In addition, Fig. \ref{fig:results_small_stud} shows how the attention maps (and therefore the translated outputs images) change after applying the proposed attention distillation loss. 
After the training with the attention loss, the attention maps in the student network are in fact less noisy and they segment and identify the target domains in the image almost perfectly. 

\noindent\textbf{Quantitative Results. }In order to further prove the effectiveness of our method we also performed a quantitative evaluation of these results. First of all, we fine-tuned a Resnet-50 classifier pre-trained on ImageNet \cite{imagenet_cvpr09} on the 7 target domains used in our experiments and we then ran it over the generated samples from our system getting a classification accuracy. Then, we calculated the FID score \cite{heusel2017gans} (the lower the better) for the same samples. The results of these quantitative evaluation steps are given in Table \ref{tab:class_stud}. 

Looking at the classification results, the student model trained with attention distillation loss outperforms the one trained without attention distillation loss proving the effectiveness of our approach. Nevertheless, the classification results are misleading in the case of gender-related attributes like \textit{goatee} or \textit{mustache}. For example, since there are no women with goatee in the real image dataset, the classification network learns to link attribute and gender therefore penalizing our system that is able to apply the target domain without changing the gender of the input person. This observation is also validated in Fig.~\ref{fig:qualitative_res}.  Regardless, transferring the attention without any other additional information like features or classification score has shown to be enough to improve the performance of a weaker network, demonstrating that visual explanation serves more purpose than just visualization and can be used during training with success. 


Regarding the FID score, the student model trained with attention distillation loss shows a slight boost in visual quality over the one trained without attention distillation loss but with a superior ability in translating the domains. 

\subsection{Pseudo-Attention Knowledge Transfer}
In this test, the objective is to use ``pseudo"-attentions defined in Sec. \ref{section3_4} to transfer knowledge between a teacher \textbf{\textit{T}} network targeting on a set of domains and a smaller student \textbf{\textit{S}} network targeting on a different set of domains. Firstly, we trained one teacher model ($\boldsymbol{\mathit{T}}^{pse}$) to translate input images to 4 facial attributes (indicated as \textbf{\textit{c}}$^{pse}$) that are \textit{\textbf{black hair}}, \textit{\textbf{blonde hair}}, \textit{\textbf{wearing hat}} and \textit{\textbf{wavy hair}}. Next, we trained the student network \textbf{\textit{S}} to translate input images to 4 different facial attributes (indicated as \textbf{\textit{c}}) that are \textit{\textbf{brown hair}}, \textit{\textbf{gray hair}}, \textit{\textbf{bald}} and \textit{\textbf{straight hair}} with the attention distillation loss. Finally, ``pseudo"-attentions (introduced in Sec. \ref{section3_1}) were used to to transfer knowledge between the teacher network and the student network. 

\begin{figure}[ht]
\begin{center}
\includegraphics[width=\linewidth]{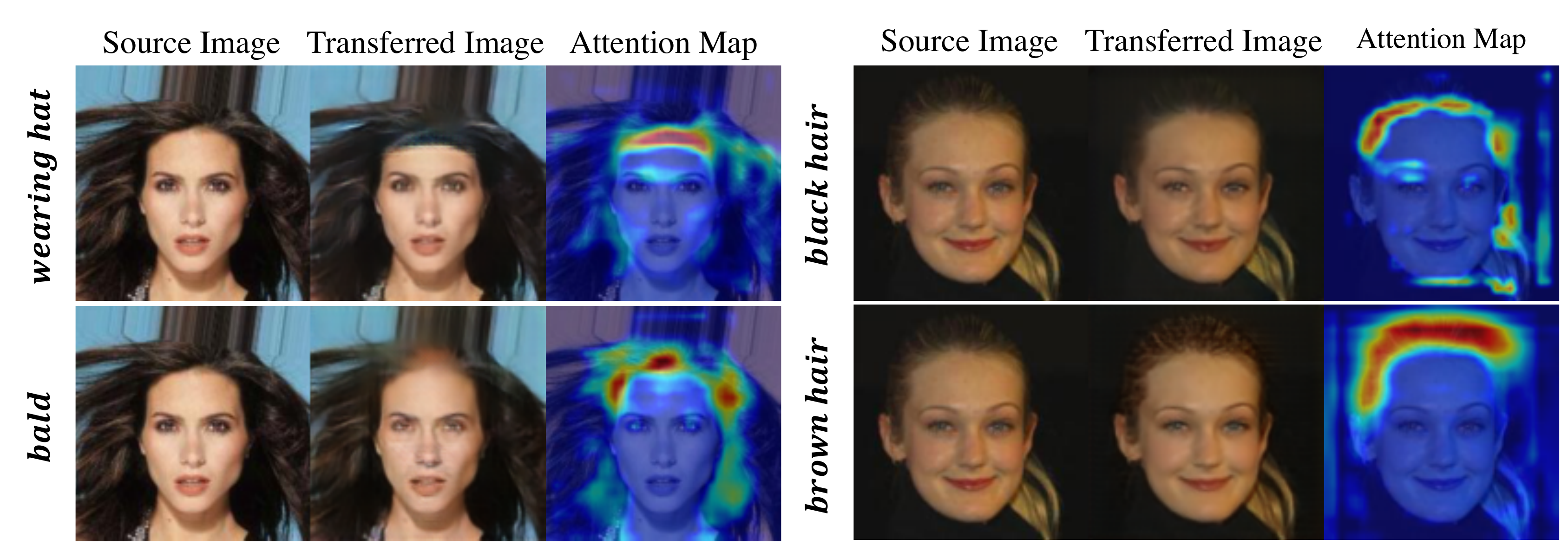}
\end{center}
\vspace{-2mm}
\caption{``Pseudo"-attention maps generated from networks targeting on two different sets of domains. From top to down, attributes in first column are: \textit{wearing hat}, \textit{bald}; Attributes in second column are: \textit{black hair}, \textit{brown hair}.}
\label{fig:results_pseudo_att_pse_tar}
\end{figure}

Attention maps obtained from the teacher models ($\boldsymbol{\mathit{T}}^{pse}$ and $\boldsymbol{\mathit{T}}$) are shown in Fig. \ref{fig:results_pseudo_att_pse_tar}. On the top are attention maps generated from the teacher model trained on a set of domains \textbf{\textit{c}}$^{pse}$ (\textit{wearing hat} and \textit{black hair}), while on the bottom are attention maps generated from another teacher model trained on a set of domains \textbf{\textit{c}} (\textit{bald} and \textit{brown hair}).
From each column including triplet images, we can see that attention maps (last column) of each pair of domains (\textit{wearing hat} vs. \textit{bald} and \textit{black hair} vs. \textit{brown hair})
share some common regions in the input images and this provides a strong evidence that it is possible to transfer knowledge by defining and using ``pseudo"-attentions from a teacher network to a student network learning a new set of domains. 


\begin{table*}[th]
    \centering
    \resizebox{\linewidth}{!}{
    \begin{tabular}{|c|c|c|c|c|c|c|c||c|c|}
    \hline
        & Blonde H. $\uparrow$ & Mustache $\uparrow$ & Goatee $\uparrow$ & Eyeglasses $\uparrow$ & Rosy Cheeks $\uparrow$ & Pale Skin $\uparrow$ & H. Makeup $\uparrow$ & Mean $\uparrow$ & FID $\downarrow$ \\ \hline
        \makecell{\textbf{\textit{S-Lite}} w/o att loss} &  58.13\% &  31.74\% &  \textbf{64.17\%} &  45.46\% &  10.23\% &  61.95\% &  94.02\% &  52.24\% &  \textbf{42.42} \\ \hline
        \textbf{\makecell{\textit{S-Lite} with att loss (Ours)} }&  \textbf{58.57\%} &  \textbf{31.93\%} &  59.35\% &  \textbf{49.78\%} &  \textbf{18.03\%} &  \textbf{72.88\%} &  \textbf{94.74\%} &  \textbf{55.04\%} &   42.51\\ \hline
    \end{tabular}
    }
    \caption{Classification results of \textbf{\textit{S-Lite}} models trained with or without attention distill loss.} 
    \label{tab:class_stud_very_light}
\end{table*}

\begin{figure}[ht]
\begin{center}
\includegraphics[width=\linewidth]{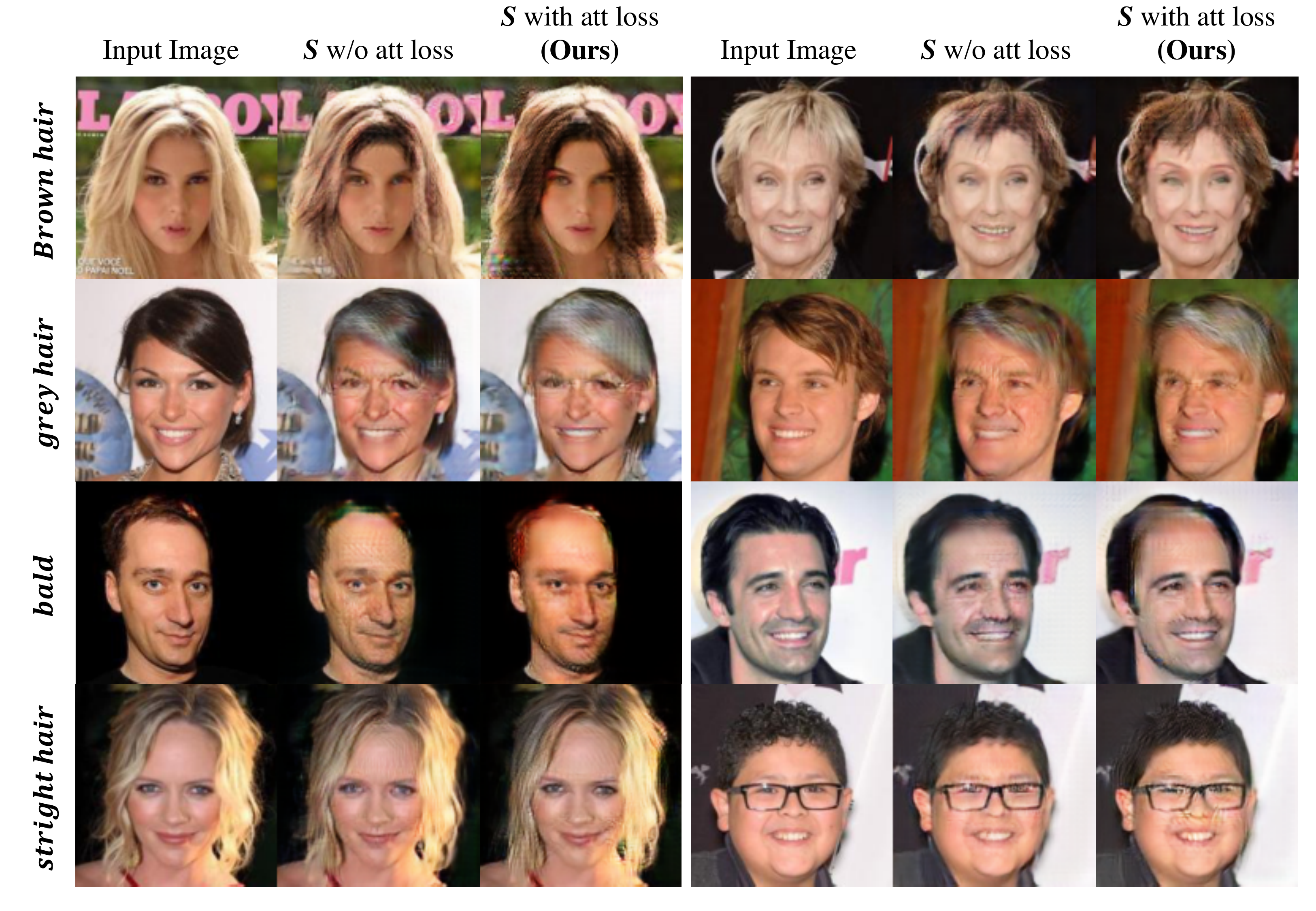}
\end{center}
\vspace{-2mm}
  \caption{Collection of comparisons between results obtained from the student \textit{\textbf{S}} trained with and without the ``pseudo"-attention distillation loss.} 
\label{fig:results_pseudo_att}
\end{figure}

\noindent\textbf{Qualitative Results.}
Next, we present transferred images in Fig. \ref{fig:results_pseudo_att}. We can see that results of the student network trained with the attention distillation loss are more convincing than the outputs from the student network that does not employ attention distillation loss. This is particularly obvious for facial attributes of \textit{brown hair}, \textit{grey hair} and \textit{bald}. For instance, in the second row, given the input image and facial attribute \textit{``grey hair"} as the target domain, the output face images (the third image in the triplet) are synthesised much better in the area \textit{``hair"} colored with \textit{``grey"}.

\begin{figure}[ht]
\begin{center}
\includegraphics[width=\linewidth]{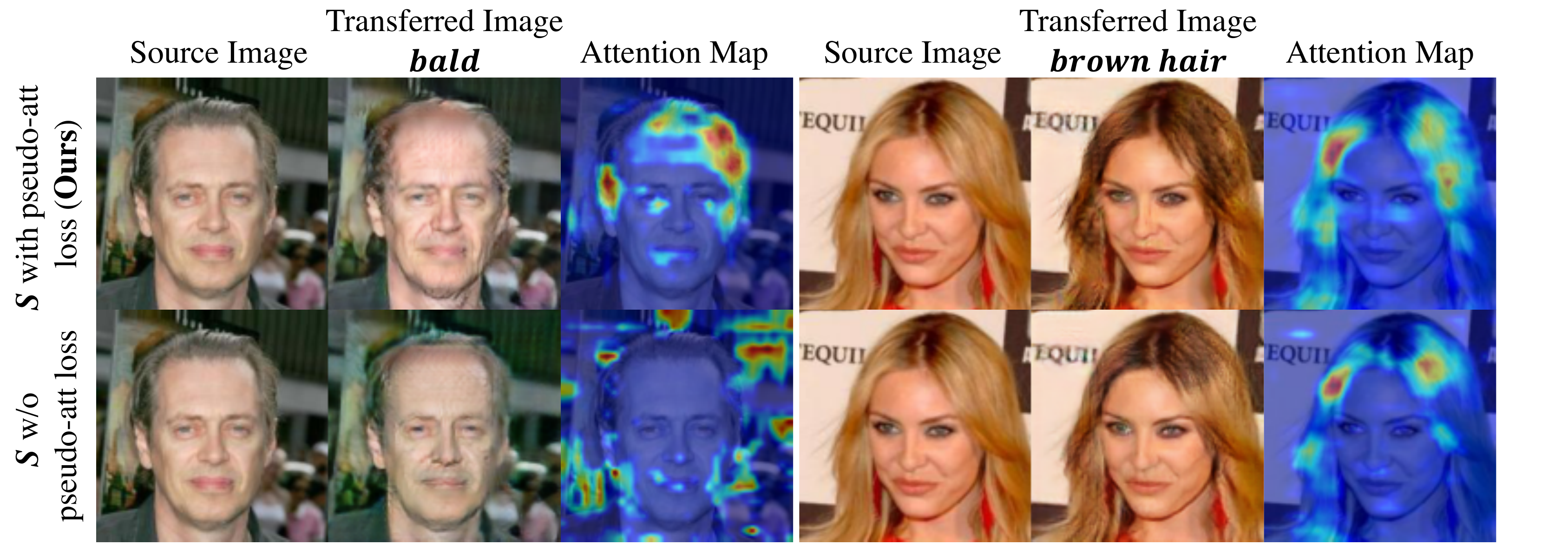}
\end{center}
\vspace{-2mm}
  \caption{Results (transferred images and attention maps in each triplet) without attention distillation loss (top) and with attention distillation loss using ``pseudo"-attention (bottom) generated from student \textbf{\textit{S}} network.} 
\label{fig:attention_maps_pseudo_att}
\vspace{-4mm}
\end{figure}

In addition to that, we show how the attention maps change after applying the attention distillation loss with ``pseudo"-attentions in Fig. \ref{fig:attention_maps_pseudo_att}. We can see that, using the pseudo-attention distillation loss, the generated image are better than the ones obtained without the pseudo-attention distillation loss. Moreover, the attention maps in the student network can indicate important regions corresponding to the target domains applied in the images. Specifically, 
the highlighted region in the attention map when the attention distillation loss is applied (the last image of the triplet) covers more area related to the attribute \textit{hair} than the attention map obtained without using attention distillation loss.

\noindent\textbf{Quantitative Results.} Furthermore, following the quantitative evaluation experiment in Sec. \ref{section4_1}, we also fine-tuned a classifier 
on the 4 target domains \textbf{\textit{c}} used in our experiments. From results in Table \ref{tab:class_stud_pseudo_att}, the student network trained with attention distillation loss using ``pseudo"-attentions outperforms the network trained without it, which proves the effectiveness of our approach.
\vspace{-2mm}
\begin{table}[h]
    \centering
    \resizebox{0.5\textwidth}{!}{
    \begin{tabular}{|c|c|c|c|c||c||c|}
    \hline
    & Brown Hair $\uparrow$ & Gray Hair $\uparrow$ & Bald $\uparrow$ & Straight Hair $\uparrow$ & Mean $\uparrow$ & FID $\downarrow$\\
    \hline
    \large \textbf{\makecell{\textit{S} w/o \\  att loss}} & \Large 79.16\% & \Large 32.46\% & \Large 8.03\% & \Large 56.48\% & \Large 44.03\% & \Large 27.02 \\
    \hline
    \large \textbf{\makecell{\textit{S} with \\ att loss (Ours)}} & \Large \textbf{82.77\%} & \Large \textbf{32.66\%} & \Large \textbf{12.43\%} & \Large \textbf{86.13\%} & \Large \textbf{53.5\%} & \Large \textbf{24.72} \\ 
    \hline
    \end{tabular}
    }
    \caption{Classification results and FID scores over 4 different facial attributes using ``pseudo"-attention maps.}
    \label{tab:class_stud_pseudo_att}
\end{table}

\vspace{-2mm}
\subsection{Ablation Experiments}
\label{sec_4_3}

\noindent \textbf{Experiments with \textit{S-Lite} Network.} Firstly, to further prove the efficacy of the proposed attention distillation loss in teacher-student paradigm trained on CelebA \cite{liu2015faceattributes} dataset, we designed an extremely light student network \textbf{\textit{S-Lite}}, which has only 1 residual block and half of the feature map of the student \textbf{\textit{S}} in each layer. 
The classification results of generated images for 7 facial attributes can be seen in Table \ref{tab:class_stud_very_light}. Even if this was a very difficult setting, looking at the classification results, we still managed to improve the overall results of the student \textit{\textbf{S-Lite}} network and the same considerations made in Sec. \ref{section4_1} are valid here. 


\noindent \textbf{Experiments on Human Facial Expression Dataset.} Secondly, we experimented on another dataset-the Radboud Faces Database (RaFD) \cite{langner2010presentation} which is a dataset consisting of different human expressions. We employ extremely light student network \textbf{\textit{S-Lite}} in our teacher-student training design because we empirically observed that our student network was already very good at solving the human expression synthesis problem. 


\begin{table}[h]
    \centering
    \resizebox{0.5\textwidth}{!}{
    \begin{tabular}{|c|c|c|c|c||c|}
    \hline
         & Disgusted $\uparrow$ & Fearful $\uparrow$ & Happy $\uparrow$ & Sad $\uparrow$ & Mean $\uparrow$\\
         \hline
         StarGAN \cite{choi2018stargan} (upper bound) & 100.00\% & 98.95\% & 100.00\% & 93.75\% & 98.17\% \\
         \hline
         \hline
         \textbf{\textit{S-Lite}} w/o att loss & 100.00\% & 95.31\% & 96.85\% & 70.57\% & 90.68\% \\
         \hline
         \textbf{\textbf{\textit{S-Lite}} with att loss (Ours)} & \textbf{100.00\%} & \textbf{97.13\%} & \textbf{97.91\%} & \textbf{72.13\%} & \textbf{91.79\%}  \\ 
         \hline
    \end{tabular}
    }
    \caption{Comparisons of classification results over 4 different human expressions.} 
    \label{tab:rafd_stud_att}
\end{table}

Quantitative results are presented in Table \ref{tab:rafd_stud_att}. Here FID score was not calculated since the test images in the dataset were too few ($\sim$1.5k) to obtain a meaningful result. 

\section{Summary and Future Work}

In this paper, we discussed the generation of visual explanation for multi-domain image-to-image conditional generative adversarial network and show how they can be used during training to improve the generated results. In particular, firstly we developed a system where a teacher transfers its attention to a smaller student using an attention distillation loss in order to make the student produce better samples. Secondly, We tested the approach on facial attributes transfer and proved its effectiveness. Thirdly, we also experimented with a setup in which the teacher and the student were trained on different, but similar in shape, domains and used ``pseudo"-attention obtained from the teacher to improve the training of the student. Finally, we tested our system on human expression synthesis showing promising results.

Future work for this project mainly includes further improving the attention generation and testing our system on a broader range of architectures.

\section{Acknowledgement}
This work was supported partially by NSF grant 1911197  and Bourns Endowment funds.

{\small
\bibliographystyle{ieee_fullname}
\bibliography{egbib}
}

\end{document}